\documentclass[letterpaper]{article} 
\usepackage[preprint]{aaai2027} 
\usepackage[hyphens]{url}  
\usepackage{graphicx} 
\usepackage{natbib} 
\usepackage{caption} 
\usepackage{amsmath}
\usepackage{amssymb} 
\usepackage{algorithm}
\usepackage{algorithmic}

\usepackage{newfloat}
\usepackage{listings}
\DeclareCaptionStyle{ruled}{labelfont=normalfont,labelsep=colon,strut=off} 
\floatstyle{ruled}
\newfloat{listing}{tb}{lst}{}
\floatname{listing}{Listing}

\usepackage{booktabs}

\title{A Visual Dependence-Aware Framework for Multimodal Unsupervised Continual Post-Training}
\author{
    Kaichen Li\textsuperscript{\rm 1,\rm 2,\rm 3},
    Zhilin Zhu\textsuperscript{\rm 4},
    Jianhao Huang\textsuperscript{\rm 1,\rm 2,\rm 3},
    Zhengqin Lai\textsuperscript{\rm 4},
    Baochen Xiong\textsuperscript{\rm 1,\rm 2,\rm 3},\\
    Zibo Shao\textsuperscript{\rm 1,\rm 2,\rm 3},
    Yaguang Song\textsuperscript{\rm 2},
    Linhui Xiao\textsuperscript{\rm 2},
    Xiaoshan Yang\textsuperscript{\rm 1,\rm 2,\rm 3},
    Changsheng Xu\textsuperscript{\rm 1,\rm 2,\rm 3}
}

\affiliations{
    \textsuperscript{\rm 1}State Key Laboratory of Multimodal Artificial Intelligence Systems, Institute of Automation,\\
    Chinese Academy of Sciences, Beijing, China\\
    \textsuperscript{\rm 2}Pengcheng Laboratory, Shenzhen, China\\
    \textsuperscript{\rm 3}School of Artificial Intelligence, University of Chinese Academy of Sciences, Beijing, China\\
    \textsuperscript{\rm 4}Harbin Institute of Technology, Shenzhen, China
}

\begin{document}

\maketitle

\begin{abstract}
In this paper, we explore a novel task of Multimodal Unsupervised Continual Post-Training (MU-CPT), enabling deployed MLLMs to continually evolve from streaming unlabeled data. Existing unsupervised post-training methods for MLLMs typically optimize target tokens uniformly, overlooking their heterogeneous visual dependence (VD). However, we reveal that token-level VD is crucial for MU-CPT. Specifically, its structural distortion serves as an indicator of cross-modal catastrophic forgetting and its inherent heterogeneity acts as a compass to guide new-task learning. Leveraging this property, we propose a Visual Dependence-Aware (VDA) framework with two main components. First, Visually Constrained Optimal Transport (VC-OT) formulates the VD structural distortion of old-task VD during new-task learning as an optimal transport problem to mitigate cross-modal forgetting. By designing a region-aware ground cost and a dependence-stratified transport penalty, it prevents global shifts in visual focus, while strictly prohibiting visual reliance from degenerating into language bias. Second, Visually Modulated Adaptation (VMA) exploits VD heterogeneity to emphasize visually grounded new-task learning, promoting new-task plasticity. Together, our method simultaneously maintains old-task stability and new-task plasticity during challenging MU-CPT. Extensive experiments under our MU-CPT setting validate the effectiveness of VDA. 
\end{abstract}

\section{Introduction}
Multimodal Large Language Models (MLLMs)~\cite{li2024llava,bai2025qwen25vltechnicalreport,bai2025qwen3,wang2025internvl35advancingopensourcemultimodal} have achieved remarkable progress by integrating visual perception with language understanding, enabling them to handle complex multimodal inputs and instructions. 
To further improve MLLMs' performance on specific tasks or user demands, most existing approaches, such as supervised fine-tuning (SFT) and reinforcement learning (RL), still largely rely on carefully curated and human-annotated multimodal data. 
As the scope and diversity of real-world applications continue to expand, empowering MLLMs through large-scale human annotation becomes prohibitively expensive and difficult to sustain. 
This has motivated growing interest in unsupervised post-training for MLLMs~\cite{zhou2024calibrated,wei2026first}, which aims to \textit{leverage unlabeled multimodal data} for model improvement. 

\begin{figure}[!t]
    \centering
    \includegraphics[width=\columnwidth]{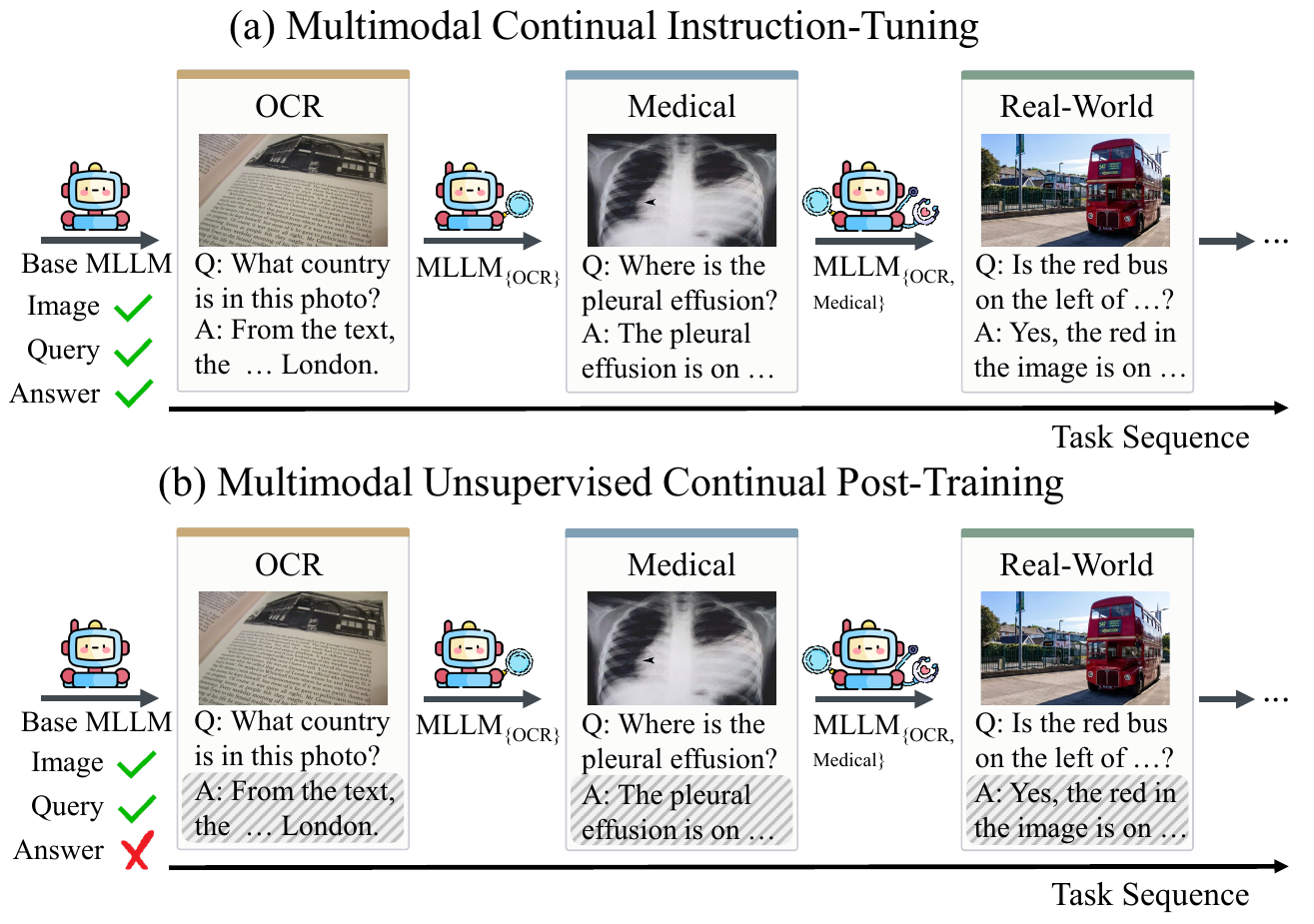}
    \caption{ 
    Comparison between (a) Multimodal Continual Instruction-Tuning and (b) Multimodal Unsupervised Continual Post-Training (MU-CPT), where labeled  answers are completely unavailable during training. 
    }
    \label{fig1}
\end{figure}

Although these unsupervised approaches have made important progress~\cite{zhu2024self,singh2026ttrv}, they are primarily developed under static, closed-world settings, where unlabeled data are collected in advance from a predefined distribution that remains unchanged~\cite{zhu2024reshaping,zhu2026dance}. However, real-world applications of MLLMs are inherently dynamic~\cite{zhou2024calibrated,zhu2024self,xu2026evomas}. 
MLLMs deployed in open environments continuously encounter evolving scenarios~\cite{zhu2026sample}, yielding non-stationary streams of unlabeled multimodal data.  
Ideally, MLLMs should continually strengthen their capabilities in emerging scenarios while preserving those learned from earlier experience~\cite{shao2026pivotmerge,shao2026agentpatch}. 
However, existing unsupervised post-training methods~\cite{hu2025test} struggle with non-stationary data distributions. 
Meanwhile, existing multimodal continual instruction-tuning (MCIT)~\cite{chen2024coin,guo2025hide} approaches assume access to abundant and expensive labeled multimodal data, as illustrated in Fig.~\ref{fig1}(a).  
Bridging these two paradigms is therefore essential to enable sustained evolution of MLLMs through unlabeled multimodal data streams. 
To bridge this gap, we formalize a novel setting termed Multimodal Unsupervised Continual Post-Training (MU-CPT). 
As illustrated in Fig.~\ref{fig1}(b), under MU-CPT setting, MLLMs sequentially encounter distinct multimodal tasks without access to the corresponding answers during continual learning. The model is expected to \textit{acquire} capabilities for each new task and \textit{preserve} those learned from previous tasks. 

\begin{figure}[!t]
    \centering
    \includegraphics[width=\columnwidth]{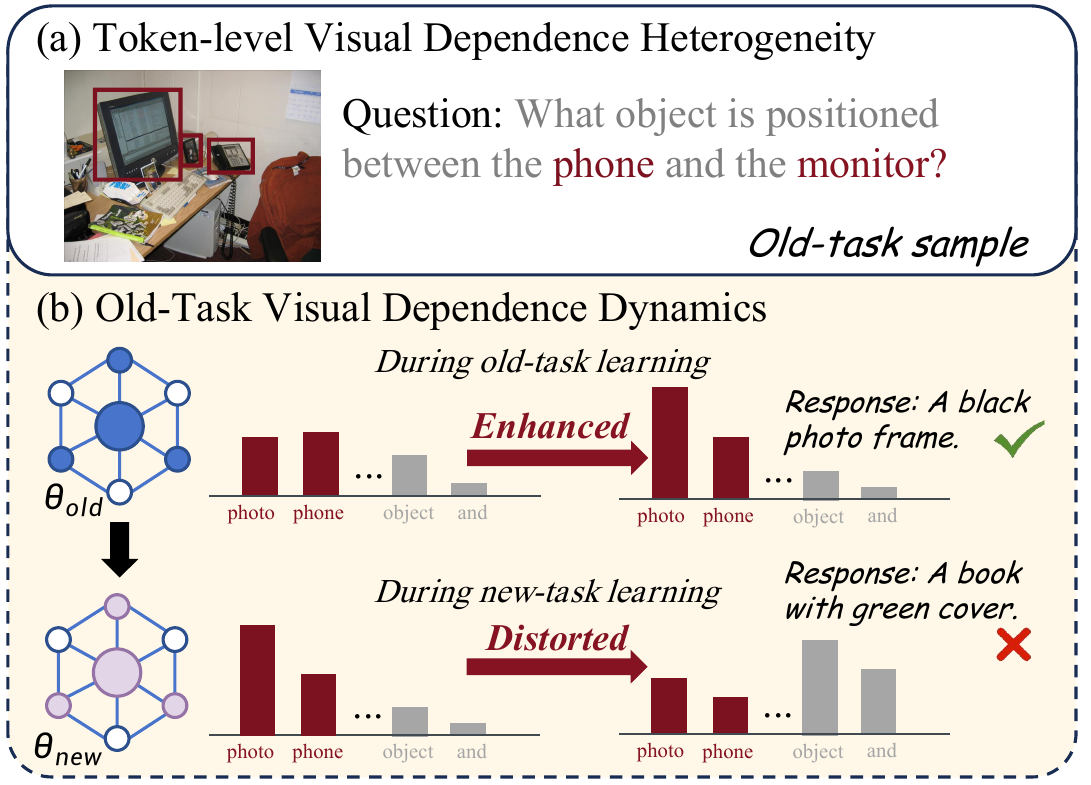}
    \caption{
(a) Text tokens in multimodal data exhibit \textbf{\textit{heterogeneous levels of visual dependence}}, where red and grey denote high and low visual dependence, respectively. 
(b) Learning new tasks leads to \textbf{\textit{structural distortion}} of visual dependence on old tasks.
    }
    \label{fig2}
\end{figure}

Existing unsupervised post-training research has primarily concentrated on how to construct optimization target sequences~\cite{zhu2024self,wei2026first,yu2026stabilizing}. However, these approaches invariably treat the constructed target sequences as monolithic entities, implicitly optimizing all target tokens uniformly, thereby leaving a fine-grained property intrinsic to the learning signals themselves largely unexamined: token-level Visual Dependence (VD) heterogeneity~\cite{ye2026not}. 
By measuring how much real visual input increases a token’s log-likelihood relative to an information-free counterfactual visual input, we observe that text tokens in multi-modal data naturally bifurcate into two regimes (see Figure~\ref{fig2} (a)): visually grounded entities exhibit pronounced VD, whereas language-driven contextual markers exhibit negligible VD. 
Crucially, as shown in Figure~\ref{fig2}(b), we further identify that sequential unsupervised adaptation induces a severe failure mode which we formalize as {cross-modal attributional forgetting}: the once-concentrated VD on visually grounded tokens (e.g., \textit{"photo"}, \textit{"phone"}) is progressively flattened and redistributed toward language-driven tokens (e.g., \textit{"object"}, \textit{"and"}), and this structural drift co-occurs with the model abandoning the correct, visually grounded answer for an hallucinated one. 

Motivated by these observations, we propose a novel Visual Dependence-Aware (VDA) framework, leveraging token-level VD as a unified basis to harmonize old-task stability and new-task plasticity, a fundamental challenge in continual learning~\cite{parisi2018continual,de2021continual}. Our framework comprises two synergistic components. First, we innovatively formulate the mitigation of cross-modal attributional forgetting as a Visually Constrained Optimal Transport (VC-OT) problem, which jointly preserves the overall strength and the structural organization of visual dependencies. Specifically, we explicitly structure the optimal transport matrix through two novel mechanisms: a region-aware cost that regulates internal VD redistribution based on visual spatial similarity, and a cross-set penalty that strictly blocks the erroneous transport of visual reliance toward vision-independent tokens. Beyond preserving old-task visual dependencies, effective continual learning also requires sufficient plasticity to acquire new capabilities. Accordingly, we further introduce Visually Modulated Adaptation (VMA), which promotes new-task learning by emphasizing visually grounded tokens while attenuating task-specific language patterns that may interfere with previous tasks. Consequently, these designs enable our framework to maintain strong stability without compromising plasticity throughout the MU-CPT process. 

Our contributions are summarized as follows: 
(1) We formalize Multimodal Unsupervised Continual Post-Training for MLLMs (MU-CPT), a novel and practical setting that enables MLLMs to continuously evolve from non-stationary, unlabeled multimodal data streams.
(2) We uncover the phenomenon of cross-modal attributional forgetting caused by token-level visual dependence distortion, establishing VD as a unified signal to balance stability and plasticity.
(3) We propose the {VDA} framework, introducing {Visually Constrained Optimal Transport} to preserve structural visual grounding and {Visually Modulated Adaptation} to mitigate linguistic bias. 
(4) Comprehensive experiments across diverse multimodal tasks validate the effectiveness of our method, setting a strong baseline for unsupervised MLLM evolution.

\section{Related Work}
\textbf{Continual Learning.}
Continual learning (CL) seeks to balance new-task plasticity and old-task stability, with representative solutions based on regularization~\cite{kirkpatrick2017overcoming,li2017learning}, rehearsal~\cite{rebuffi2017icarl,lopez2017gradient}, and parameter isolation~\cite{mallya2017packnet}; continual self-supervised learning further extends CL to unlabeled streams, mainly for unimodal representation learning~\cite{madaan2021representational}. Multimodal CL subsequently studies sequential VQA and prompt-based adaptation~\cite{zhang2023vqacl,qian2023decouple}. More recently, continual instruction tuning (CIT) extends CL to instruction-following MLLMs trained sequentially on labeled image--instruction--answer data, with EProj and CoIN establishing representative benchmarks~\cite{he2026continual,chen2024coin}. Existing CIT methods mitigate forgetting through selective attention distillation in SEEKR-MLLM~\cite{he2024seekr}, global--local expert adaptation in CL-MoE~\cite{huai2025cl}, old-task-important LoRA regularization in SEFE~\cite{chen2025sefe}, or replay-augmented gradient guidance in DGG~\cite{li2025multimodal}. However, these methods rely on labeled answers to define learning and preservation targets, whereas MU-CPT must balance plasticity and stability using only answer-unlabeled multimodal streams.

\noindent\textbf{Unsupervised Post-Training.}
Unsupervised post-training adapts models to unlabeled inputs using self-generated learning signals. For LLMs, LSMI selects high-confidence solutions through self-consistency~\cite{huang2023large}, ScPO constructs preferences from response consistency~\cite{prasad2024self}, and TLM minimizes input perplexity~\cite{hu2025test}. Related studies also exploit semantic or token-level entropy~\cite{zhang2026right,agarwal2026unreasonable}. For MLLMs, SeVa contrasts responses generated from original and perturbed images~\cite{zhu2024self}, CSR calibrates self-rewards using image--response relevance~\cite{zhou2024calibrated}, MM-UPT derives pseudo-rewards through majority voting~\cite{wei2026first}, and TTRV constructs rewards from response frequency and output entropy~\cite{singh2026ttrv}; CSRS further stabilizes self-rewarding through retraced sampling and softened frequency rewards~\cite{yu2026stabilizing}. Although these methods provide diverse unsupervised objectives, they mainly assume static unlabeled datasets and overlook both continual forgetting and token-level VD heterogeneity.

\section{Methodology}
\subsection{Problem Formulation}
We formulate Multimodal Unsupervised Continual Post-Training (MU-CPT), enabling models to continually evolve on unlabeled multimodal data while preserving previously acquired capabilities, unlike standard continual learning that relies on fully annotated data. 
Formally, under this setting, we consider an instruction-tuned MLLM, initially parameterized by $\theta_0$, continually learning from $K$ multimodal tasks. 
At stage $k$, the model only accesses an answer-unlabeled multimodal data set $\mathcal{D}_k = \{(I_k^i, Q_k^i)\}_{i=1}^{N_k}$, where $I_k^i$ and $Q_k^i$ denote the $i$-th image and its associated question, respectively. 
The overall objective is to acquire capabilities for the current task from $\mathcal{D}_k$ while preventing catastrophic forgetting on previously learned tasks, yielding the updated parameters $\theta_k$.
Upon completing stage $k$, the updated model $\theta_k$ is evaluated on held-out image-question pairs from all tasks encountered so far. 
Regardless of the specific form of the unsupervised signal, we denote the optimization target sequence of length $L$ for an $(I,Q)$ pair as $Y=(y_1,\ldots,y_L)$, where $t\in\{1,\ldots,L\}$ indexes the target-token position, and let $\ell_t(\theta)$ denote the loss associated with token $y_t$. 

\subsection{Framework Overview}
\begin{figure*}[t]
    \centering
    \includegraphics[width=0.95\textwidth]{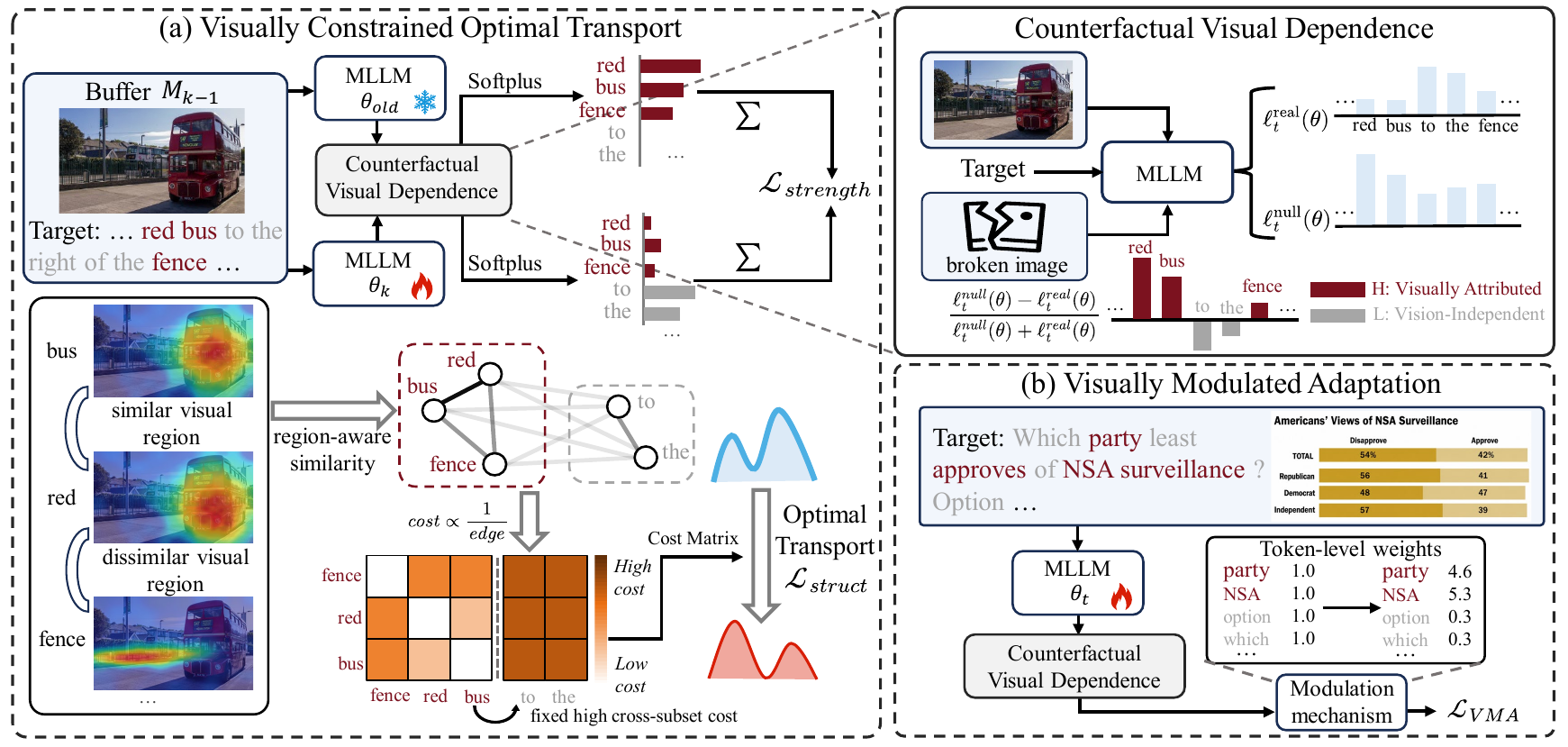}
    \caption{
Overview of the proposed Visual Dependence-Aware (VDA) framework. (a) Visually Constrained Optimal Transport (VC-OT) preserves old-task VD strength and structure through region-aware and dependence-stratified transport. (b) Visually Modulated Adaptation (VMA) reweights incoming-task tokens according to their VD to promote visually grounded learning. Together, they effectively balance stability and plasticity in MU-CPT. 
}
    \label{fig:framework}
\end{figure*}

While existing unsupervised methods mainly construct sequence-level targets, they typically optimize all target tokens uniformly, overlooking token-level visual dependence (VD) heterogeneity. In continual learning, this variation provides an intrinsic anchor for balancing stability and plasticity. Accordingly, VDA addresses MU-CPT at the token level rather than only at the sequence level, as illustrated in Figure~\ref{fig:framework}. 

To maintain old-task stability, Visually Constrained Optimal Transport (VC-OT) formulates old-task VD distortion as an optimal transport problem. Its region-aware cost and dependence-stratified penalty suppress global shifts in visual focus and transfer toward vision-independent tokens. To preserve new-task plasticity, Visually Modulated Adaptation (VMA) emphasizes visually grounded tokens while attenuating task-specific language patterns that cause cross-task interference. Together, these complementary mechanisms enable robust continual learning from unlabeled multimodal data.

\subsection{Token-Level Visual Dependence}
To quantify the visual dependence of each target token $y_t \in Y$, we compare its prediction under real and counterfactual visual conditions. Specifically, keeping the target token and its preceding textual context unchanged, we compute:
\begin{equation}
\begin{aligned}
    \ell_t^{\mathrm{real}}(\theta)
    &=
    -\log p_{\theta}
    \left(
        y_t \mid I,\mathcal{P},y_{<t}
    \right), \\
    \ell_t^{\mathrm{null}}(\theta)
    &=
    -\log p_{\theta}
    \left(
        y_t \mid I_{\mathrm{null}},\mathcal{P},y_{<t}
    \right),
\end{aligned}
\label{eq:vd_losses}
\end{equation}
where $I_{\mathrm{null}}$ denotes a counterfactual visual input with no information. (See Appendix for details) The token-level visual dependence is then defined as:
\begin{equation}
    D_t(\theta)
    =
    \frac{
        \ell_t^{\mathrm{null}}(\theta)
        -
        \ell_t^{\mathrm{real}}(\theta)
    }{
        \ell_t^{\mathrm{null}}(\theta)
        +
        \ell_t^{\mathrm{real}}(\theta)
    }.
    \label{eq:visual_dependence}
\end{equation}
A positive $D_t$ indicates that the visual input actively facilitates the prediction of the token, whereas a non-positive value suggests that the token is driven primarily by language context. 

During continual learning, this token-level metric serves a dual purpose by explicitly bridging visual dependence with the model's core cross-modal attribution capability. For replayed old-task samples, the structural distortion of VD reflects the collapse of this capability, marking a severe cross-modal attributional forgetting that necessitates VD structural preservation to maintain stability. On the other hand, while pursuing stability alone risks suppressing model adaptability, VD heterogeneity acts as an intrinsic compass to locate visually informative tokens. By directing the optimization focus toward these essential visual cues rather than task-specific language biases, it effectively fosters new-task plasticity. This dual functionality establishes VD as a unified foundation for our subsequent designs.

\subsection{Visually Constrained Optimal Transport}
Preserving old-task VD is essential to prevent cross-modal attributional forgetting. To this end, we formulate the structural drift between the reference and current old-task VD as a unified optimal transport problem. By leveraging the highly non-uniform damage that VD redistribution in different directions inflicts on the model's cross-modal attribution capability, we introduce a region-aware transport cost and dependence-stratified transport penalty to safely regulate internal VD structural shifts. Furthermore, we explicitly constrain the overall VD strength to prevent the global collapse of visual reliance. 

\paragraph{VD Transport Formulation.}
Since only $D_t > 0$ signifies that visual evidence actively supports token generation, we focus exclusively on positive visual dependence. To extract this non-negative mass for optimal transport while preserving smooth gradients, we smoothly approximate the hard $\max(\cdot, 0)$ truncation with a low-temperature Softplus, projecting VD into nonnegative mass:
\begin{equation}
    r_t^{m}
    =
    \tau_D
    \operatorname{Softplus}
    \left(
        \frac{D_t^{m}}{\tau_D}
    \right),
    \qquad
    m\in\{\mathrm{ref},\mathrm{cur}\}.
    \label{eq:positive_vd_mass}
\end{equation}
Here, a size-bounded buffer retains a small subset of old-task samples and their reference VD. For each replayed sample, $D_t^{\mathrm{ref}}$ is computed by the model immediately after learning its source task and cached in the buffer, while $D_t^{\mathrm{cur}}$ is computed by the current trainable model. 
While this absolute mass $r_t^{m}$ will be explicitly constrained later to maintain the overall VD strength, modeling its internal structural flow requires a relative mass distribution. Therefore, we normalize $r_t^{m}$ within the target sequence: 
\begin{equation}
    \tilde{r}_t^{m}
    =
    \frac{r_t^{m}}{\sum_{u=1}^{L}r_u^{m} },
    \qquad
    m\in\{\mathrm{ref},\mathrm{cur}\}.
    \label{eq:normalized_r}
\end{equation}
where $L$ represents the length of the target sequence. 
We treat $\tilde{\mathbf{r}}^{\mathrm{ref}}$ as the source distribution
and $\tilde{\mathbf{r}}^{\mathrm{cur}}$ as the target distribution.
Specifically, $T_{ij}$ denotes the VD mass transported from source token
$i$ under the reference model to destination token $j$ under the current
model, satisfying
$\sum_{j=1}^{L} T_{ij} = \tilde{r}_{i}^{\mathrm{ref}}$
and
$\sum_{i=1}^{L} T_{ij} = \tilde{r}_{j}^{\mathrm{cur}}$. 
To set up the transport space, we explicitly separate tokens attributed to visual evidence from those driven by language context, partitioning the target sequence into a visually attributed set $\mathcal{H} = \{t\mid D_t^{\mathrm{ref}}>0\}$ and a vision-independent set $\mathcal{L} = \{t\mid D_t^{\mathrm{ref}}\leq0\}$. 
Given that the reference mass in $\mathcal{L}$ is negligible and does not represent any visual attribution that needs to be preserved, we focus on the structural regularization on transport originating from $\mathcal{H}$. Accordingly, we consider two relevant transport paths: intra-set redistribution within $\mathcal{H}$ and cross-set transfer from $\mathcal{H}$ to $\mathcal{L}$. 

\paragraph{Region-Aware Transport Cost.}
For the intra-set transport within $\mathcal{H}$, our primary concern is to prevent global shifts in visual focus, where the model's attention drifts entirely from one region to an unrelated area. This drift signifies a severe forgetting of how the model comprehends the image. Therefore, we permit mild redistribution among visually similar tokens while penalizing such global shifts. 
Let $\mathbf{a}_{t,\mathrm{ref}}^{\ell}$ denote the normalized attention distribution of token $t \in \{1, \dots, L\}$ over all visual tokens at decoder layer $\ell$, extracted by the model immediately after learning the sample's source task and cached with the replay sample. 
We formulate the region-aware transport cost between tokens $i$ and $j$ as:
\begin{equation}
    d_{\mathrm{reg}}(i,j)
    =
    \frac{1}{|\mathcal{G}|}
    \sum_{\ell\in\mathcal{G}}
    \operatorname{JSD}
    \left(
\mathbf{a}_{i,\mathrm{ref}}^{\ell},
\mathbf{a}_{j,\mathrm{ref}}^{\ell}
    \right).
    \label{eq:region_distance}
\end{equation}
Here, $\mathcal{G}$ specifies the selected intermediate decoder layers. $JSD \in [0,1]$ represents Jensen-Shannon Divergence, which is used to measure the discrepancy in attended regions, a greater difference directly yields a higher transport cost. 

\paragraph{Dependence-Stratified Transport Penalty.}
We next consider cross-set transport from $\mathcal{H}$ to $\mathcal{L}$.
Since tokens in $\mathcal{L}$ are vision-independent under the reference model, assigning VD mass to these positions indicates spurious visual attribution to language-driven tokens.
We therefore penalize $\mathcal{H}\!\rightarrow\!\mathcal{L}$ transport with the maximum JSD cost of 1, corresponding to completely non-overlapping visual attention distributions.

\paragraph{Structural Transport Objective.}
Combining the region-aware intra-set cost and the dependence-stratified cross-set penalty, we define the transport cost matrix as:
\begin{equation}
    C_{ij}
    =
    \begin{cases}
        d_{\mathrm{reg}}(i,j),
        & i\in\mathcal{H},\ j\in\mathcal{H}, \\[3pt]
        1,
        & i\in\mathcal{H},\ j\in\mathcal{L}, \\[3pt]
        0,
        & i\in\mathcal{L}.
    \end{cases}
    \label{eq:transport_cost}
\end{equation}
Since transport originating from $\mathcal{L}$ carries negligible reference mass and no meaningful visual attribution, we set its cost to zero.
Let $\mathbf{T}^{*}$ denote the optimal transport plan obtained through Sinkhorn iterations.
The structural alignment loss is:
\begin{equation}
    \mathcal{L}_{\mathrm{struct}}
    =
    \langle \mathbf{T}^{*}, \mathbf{C} \rangle
    =
    \sum_{i=1}^{L}
    \sum_{j=1}^{L}
    T^{*}_{ij} C_{ij}.
    \label{eq:l_struct}
\end{equation}

\paragraph{VD Strength Preservation.}
While $\mathcal{L}_{\mathrm{struct}}$ regulates the internal VD organization, learning new tasks may still reduce the overall VD strength.
We therefore match the total positive VD mass of the current model to its reference:
\begin{equation}
    \mathcal{L}_{\mathrm{strength}}
    =
    \frac{
        \left|
            \sum_{t=1}^{L}r_t^{\mathrm{cur}}
            -
            \sum_{t=1}^{L}r_t^{\mathrm{ref}}
        \right|
    }{
        \sum_{t=1}^{L}r_t^{\mathrm{ref}}
    }.
    \label{eq:vd_mass_loss}
\end{equation}
The complete VC-OT objective is:
\begin{equation}
    \mathcal{L}_{\mathrm{VC\text{-}OT}}
    =
    \mathcal{L}_{\mathrm{struct}}
    +
    \mathcal{L}_{\mathrm{strength}}.
    \label{eq:vc_ot}
\end{equation}

\subsection{Visually Modulated Adaptation}

Although VC-OT preserves old-task knowledge, anti-forgetting constraints alone can restrict new-task plasticity~\cite{jung2020continual}. To balance the two, we introduce Visually Modulated Adaptation (VMA). During new-task training, visually attributed tokens carry task-relevant visual cues, whereas vision-independent tokens often encode task-specific linguistic templates, such as multiple-choice formats. 
Overfitting to these templates can induce language bias and cross-task interference; for example, the model may generate option labels for previously learned open-ended questions. 
VMA modulates each token's learning signal according to its visual dependence, thereby emphasizing visually grounded knowledge to improve plasticity while suppressing language-bias interference. 

Let $\ell_t^{\mathrm{u}}(\theta)$ denote the unmodulated token-level loss associated with target token $y_t$. Based on its current visual dependence defined in Eq.~\ref{eq:visual_dependence}, we first compute:
\begin{equation}
\begin{aligned}
    \widetilde{w}_t
    &=
    1+\lambda_{\mathrm{VMA}}
    \operatorname{ReLU}
    \left(
        \operatorname{sg}\!\left[D_t(\theta)\right]
    \right), \\
    w_t
    &=
    \frac{
        L\widetilde{w}_t
    }{
        \sum_{u=1}^{L}\widetilde{w}_u
    },
\end{aligned}
\label{eq:VMA_weight}
\end{equation}
where $\operatorname{sg}[\cdot]$ denotes the stop-gradient operation and $\lambda_{\mathrm{VMA}}$ controls the modulation strength. The unnormalized modulation factor $\widetilde{w}_t$ amplifies the learning intensity strictly for visually attributed tokens, while retaining a base unit contribution for the rest. Sequence-level normalization ensures the average modulation scale remains neutral, avoiding sample-dependent optimization shifts.

The modulated current-task objective is then formulated as:
\begin{equation}
    \mathcal{L}_{\mathrm{VMA}}
    =
    \frac{1}{L}
    \sum_{t=1}^{L}
    w_t\,\ell_t^{\mathrm{u}}(\theta).
    \label{eq:VMA_loss}
\end{equation}
By adaptively modulating the relative contribution of visually attributed tokens, VMA encourages the model to absorb effective visual knowledge from the incoming task, while attenuating the influence of task-specific linguistic formatting. Consequently, VMA complements VC-OT by enhancing new-task plasticity without relying on external supervision. To sum up, the overall loss is given by:

\begin{equation}
    \mathcal{L}_{\mathrm{total}}
    =
    \mathcal{L}_{\mathrm{VMA}}
    +
    \mathcal{L}_{\mathrm{VC\text{-}OT}}.
    \label{eq:overall_loss}
\end{equation}

\section{Experiments}
\subsection{Experimental Setup}
\paragraph{Datasets.}
We evaluate six tasks spanning OCR, scientific diagrams, financial charts, compositional reasoning, driving scenes, and medical VQA: TextVQA~\cite{singh2019towards}, SciVQA~\cite{borisova2025scivqa}, StockQA~\cite{zhao2025mllm}, GQA~\cite{hudson2019gqa}, DriveLM~\cite{sima2024drivelm}, and PMC-VQA~\cite{zhang2023pmc}. We follow the original splits and use only image--question pairs for continual training. All methods use the same random order TextVQA$\rightarrow$SciVQA$\rightarrow$StockQA$\rightarrow$GQA$\rightarrow$DriveLM$\rightarrow$PMC-VQA (See Appendix for more random orders) and are evaluated on all seen test sets after each stage. 

\begin{table*}[t]
\centering
\small
\renewcommand{\arraystretch}{1.1}
\begin{tabular}{lccccccc}
\toprule
Method
& TextVQA
& SciVQA
& StockQA
& GQA
& DriveLM
& PMC-VQA
& AvgAcc $\uparrow$ \\
\midrule

\multicolumn{8}{l}{\textit{Frozen Backbone}} \\
Qwen2.5-VL-7B 
& 79.1
& 52.9
& 53.3
& 60.0
& 24.7
& 53.5
& 53.9 \\

\midrule
\multicolumn{8}{l}{\textit{Unsupervised Post-training Methods}} \\
LSMI (EMNLP'23)
& 79.5
& \underline{63.2}
& 62.5
& \underline{68.3}
& 29.0
& 49.7
& 58.7 \\

SeVa (ACM MM'24)
& 79.9
& 57.3
& 60.7
& 64.0
& 27.3
& 54.3
& 57.2 \\

ScPO (ICML'25)
& 80.3
& 61.7
& 64.4
& 66.0
& 28.6
& 51.6
& 58.8 \\

TLM (ICML'25)
& 77.2
& 49.3
& 43.3
& 46.1
& 29.8
& 42.4
& 48.0 \\

MM-UPT (NeurIPS'25)
& 78.8
& 54.7
& 57.8
& 62.4
& 24.8
& 55.0
& 55.6 \\

TTRV (CVPR'26)
& 79.6
& 60.5
& 59.2
& 64.4
& 26.7
& \textbf{55.9}
& 57.7 \\

\midrule
\multicolumn{8}{l}{\textit{Continual Learning Methods}} \\
SEEKR-MLLM (EMNLP'24)
& 75.4
& 61.0
& 67.8
& \textbf{68.8}
& \underline{31.7}
& \underline{55.8}
& \underline{60.1} \\

CL-MoE (CVPR'25)
& 78.5
& 60.5
& 62.5
& 64.1
& 27.7
& 54.2
& 57.9 \\

SEFE (ICML'25)
& 79.3
& 33.1
& \textbf{72.5}
& 56.8
& 26.5
& 53.1
& 53.6 \\

DGG (CVPR'26)
& \underline{84.4}
& 54.5
& 61.8
& 61.7
& 28.5
& 54.6
& 57.6 \\

\midrule
\textbf{VDA (Ours)}
& \textbf{85.9}
& \textbf{63.6}
& \underline{70.3}
& 67.6
& \textbf{32.6}
& 54.9
& \textbf{62.5} \\
\bottomrule
\end{tabular}
\caption{Comparison with existing Unsupervised Post-Training methods and Continual learning methods.}
\label{tab:main_results}
\end{table*}

\paragraph{Comparison Methods.}
We compare with six unsupervised post-training methods---LSMI~\cite{huang2023large}, SeVa~\cite{zhu2024self}, ScPO~\cite{prasad2024self}, TLM~\cite{hu2025test}, MM-UPT~\cite{wei2026first}, and TTRV~\cite{singh2026ttrv}---and four MLLM continual-learning methods---DGG~\cite{li2025multimodal}, SEFE~\cite{chen2025sefe}, CL-MoE~\cite{huai2025cl}, and SEEKR-MLLM~\cite{he2024seekr}. 

\paragraph{Implementation Details.}
We use Qwen2.5-VL-7B~\cite{bai2025qwen25vltechnicalreport} as the primary backbone (See appendix for other model families). For VDA and all continual learning baselines, we perform parameter-efficient tuning using LoRA with a rank of 128. For the unsupervised post-training baselines, we follow the training configurations provided in their official implementations. 
Following TLM~\cite{hu2025test}, we use image-conditioned question modeling as the basic unsupervised objective, where the model autoregressively predicts question tokens conditioned on the image and a task-agnostic instruction as in~\cite{zhao2024lova3}. Our VDA is not tied to this objective, with results using other unsupervised signals reported in the appendix. 
As to other hyperparameters, we set the VD smoothing temperature $\tau_D$ to $0.05$, and buffer size to 1000. The reference VD and visual attention of each replay sample is computed once when the sample is inserted into the memory and remains fixed thereafter. 
Unless otherwise specified, we use AdamW for optimization with a batch size of 1 and a learning rate of $5\times10^{-5}$. 

\paragraph{Evaluation Metrics.}
Following standard continual-learning protocols~\cite{wang2022dualprompt,smith2023coda}, AvgAcc averages final accuracy over all tasks and is our primary stability--plasticity metric. AvgF measures the average performance drop on previous tasks, while AvgLA averages each task's accuracy immediately after it is learned, diagnosing stability and plasticity, respectively. 

\begin{table}[h]
\centering
\small
\renewcommand{\arraystretch}{1.1}
\setlength{\tabcolsep}{9pt} 
\begin{tabular}{lccc}
\toprule
Method
& AvgAcc $\uparrow$
& AvgLA $\uparrow$
& AvgF $\downarrow$ \\
\midrule
SEEKR-MLLM 
& \underline{60.1}
& \underline{62.5}
& 2.9 \\

CL-MoE 
& 57.9
& 59.5
& \underline{1.8} \\

SEFE 
& 53.6
& 61.4
& 9.4 \\

DGG 
& 57.6
& 60.8
& 3.9 \\
\midrule
\textbf{VDA (Ours)}
& \textbf{62.5}
& \textbf{63.0}
& \textbf{0.6} \\
\bottomrule
\end{tabular}
\caption{Fine-grained comparison of new-task plasticity and old-task stability
with continual learning methods.} 
\label{tab:cl_results}
\end{table}

\subsection{Comparison Results}
Table~\ref{tab:main_results} reports the final performance after sequential adaptation on six multimodal tasks. 
VDA achieves the highest AvgAcc of 62.5\%, outperforming the strongest unsupervised post-training method, ScPO, by 3.7 points and the strongest continual learning baseline, SEEKR-MLLM, by 2.4 points. 
Notably, despite learning solely from answer-unlabeled multimodal data under continually shifting data distribution, VDA improves the frozen Qwen2.5-VL-7B backbone by 8.6 points, demonstrating its ability to effectively acquire and keep knowledge from non-stationary unlabeled data streams. 
Across individual tasks, existing baselines exhibit pronounced performance variation, indicating that their adaptation is sensitive to task and domain shifts under answer-free supervision. In contrast, VDA maintains consistently competitive performance across all six benchmarks, demonstrating more robust unsupervised continual learning across heterogeneous multimodal domains. 

Furthermore, we conduct a fine-grained comparison with existing supervised CL methods from the perspectives of new-task plasticity and old-task stability, where all methods are adapted to the MU-CPT setting using the same unsupervised learning signal.
As shown in Table~\ref{tab:cl_results}, VDA achieves the highest AvgLA of 63.0\% and the lowest forgetting of 0.6, together yielding the best AvgAcc of 62.5\%.
Among the baselines, SEEKR-MLLM exhibits the strongest new-task plasticity with an AvgLA of 62.5\%, but suffers substantially greater forgetting, while CL-MoE attains relatively low forgetting at the cost of a much lower AvgLA of 59.5\%.
These results demonstrate that VDA improves old-task stability without suppressing new-task adaptation, leading to a more favorable stability--plasticity balance.  

\begin{table}[h]
\centering
\small
\renewcommand{\arraystretch}{1.1}
\setlength{\tabcolsep}{4.5pt}
\begin{tabular}{ccc|ccc}
\toprule
$\mathcal{L}_{\mathrm{VMA}}$
& $\mathcal{L}_{\mathrm{strength}}$
& $\mathcal{L}_{\mathrm{struct}}$
& AvgAcc $\uparrow$
& AvgLA $\uparrow$
& AvgF $\downarrow$ \\
\midrule
&
&
& 51.0
& 61.7
& 12.8 \\

\checkmark
&
&
& 56.9
& \textbf{63.8}
& 8.2 \\

& \checkmark
&
& 58.0
& 61.5
& 4.2 \\

\checkmark
& \checkmark
&
& 60.1
& 62.9
& 3.3 \\

& \checkmark
& \checkmark
& 59.8
& 60.3
& 0.7 \\

\checkmark
& \checkmark
& \checkmark
& \textbf{62.5}
& 63.0
& \textbf{0.6} \\
\bottomrule
\end{tabular}
\caption{Ablation study of the components in VDA.}
\label{tab:ablation}
\end{table}

\paragraph{Ablation Study.}
Table~\ref{tab:ablation} evaluates the contributions of VMA, VD strength preservation, and structural transport. VMA alone increases AvgLA from 61.7 to 63.8 and AvgAcc from 51.0 to 56.9, while reducing AvgF to 8.2, indicating that VD-guided token modulation mainly improves plasticity and partially alleviates forgetting. $\mathcal{L}_{\mathrm{strength}}$ further lowers AvgF to 4.2 but slightly reduces AvgLA, reflecting the adaptation cost of stability constraints. Combining it with VMA recovers plasticity and raises AvgAcc to 60.1. Joint strength and structural preservation reduce AvgF to 0.7, but limits AvgLA to 60.3. The complete model achieves the best AvgAcc of 62.5, with AvgLA of 63.0 and AvgF of 0.6. These results show that VMA compensates for the restricted adaptation caused by VD preservation, while the two preservation terms jointly provide strong old-task stability.

\begin{figure}[h]
    \centering
    \includegraphics[width=0.9\columnwidth]{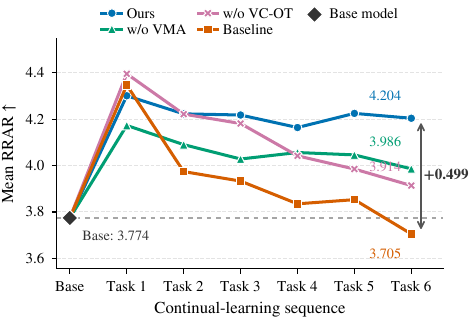}
    \caption{
    RRAR trajectories on TextVQA throughout continual-learning. 
    }
    \label{fig:rrar}
\end{figure}

\subsection{Detailed Analysis}
\paragraph{Analysis of cross-modal comprehension.}
To examine whether VDA strengthens and preserves cross-modal comprehension throughout continual learning, we track the Relevant Region Attention Ratio (RRAR)~\cite{peng2026deeperthoughtweakeraim} on TextVQA using its ground-truth bounding boxes provided by\cite{khayatkhoei2025mllms}. 
RRAR measures the relative attention allocated to question-relevant visual regions compared with the entire image.
As shown in Figure~\ref{fig:rrar}, initial adaptation increases RRAR, whereas the baseline (without both VMA and VC-OT) progressively declines as new tasks are learned, eventually falling below the frozen backbone. 
In contrast, VDA maintains an RRAR of 4.204 after the full sequence, outperforming the baseline by 0.499.
Removing VMA or VC-OT reduces the final RRAR to 3.986 and 3.914, respectively, showing that both components contribute to maintaining strong question-relevant visual attribution.
These results provide mechanism-level evidence that VDA enhances cross-modal comprehension during adaptation and prevents its degradation across subsequent tasks.

\begin{table}[h]
\centering
\small
\renewcommand{\arraystretch}{1.1}
\setlength{\tabcolsep}{2.3pt}
\begin{tabular}{@{}lcccc@{}}
\toprule
VD Protection
& \shortstack{VD Strength\\Drift (\%) $\downarrow$}
& \shortstack{VD Distribution\\Drift $\downarrow$}
& AvgAcc $\uparrow$
& AvgF $\downarrow$ \\
\midrule

No Protection
& 35.94
& 0.1052
& 56.9
& 8.2 \\

Token-Level L1
& 5.08
& 0.0187
& 59.1
& 0.4 \\

Mass+KL
& 14.89
& 0.0428
& 60.3
& 3.4 \\

\midrule
\textbf{VC-OT}
& 14.27
& 0.0385
& \textbf{62.5}
& 0.6 \\

\bottomrule
\end{tabular}
\caption{Comparison of different VD preservation strategies.
VD Strength Drift measures the relative change in total VD.
VD Distribution Drift measures the JSD between VD distributions.}
\label{tab:vd_drift}
\end{table}

\paragraph{Comparison of VD Preservation Strategies.}
We further compare VC-OT with several alternative strategies in preserving old-task VD. ``No Protection'' denotes VDA without VC-OT, while the remaining variants replace VC-OT with token-level L1 matching, or Mass+KL alignment. As shown in Table~\ref{tab:vd_drift}, No Protection exhibit substantial VD drift and severe forgetting, demonstrating the necessity of explicitly preserving old-task VD structures. Token-level L1 achieves the smallest VD drift and the lowest AvgF of 0.4, but only reaches an AvgAcc of 59.1, suggesting that rigid position-wise matching suppresses model's plasticity. Mass+KL provides a softer distribution-level constraint, yet treats all cross-token shifts uniformly, without distinguishing visually consistent redistribution from harmful transfer toward vision-independent tokens. In contrast, VC-OT achieves comparable VD preservation while using region-aware transport and a dependence-stratified transport penalty to regulate different redistribution paths, resulting in a low AvgF and the highest global AvgAcc. 

\begin{figure}[h]
    \centering
    \includegraphics[width=0.9\columnwidth]{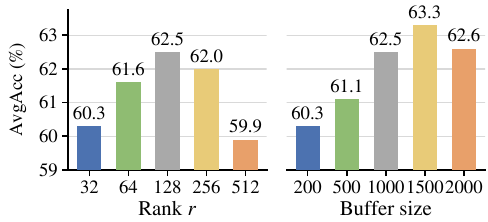}
    \caption{
    Hyperparameter Analysis.
    }
    \label{fig:hyper}
\end{figure}

\paragraph{Hyperparameter Analysis.}
Fig.~\ref{fig:hyper} analyzes the effects of the LoRA rank and memory-buffer size. Increasing the LoRA rank initially improves AvgAcc by enhancing new-task adaptation, with the best performance achieved at r=128; further increasing the rank introduces more trainable parameters, leading to greater forgetting and worse AvgAcc. Enlarging the buffer generally improves performance by providing broader coverage of previous tasks, while the gain gradually saturates. A buffer size of 1,000 already achieves a good AvgAcc of 62.5.

\section{Conclusion}
This work introduces Multimodal Unsupervised Continual Post-Training (MU-CPT), where MLLMs continually learn from non-stationary multimodal data without labeled answers. We identify token-level visual dependence (VD) as an intrinsic signal of cross-modal attributional forgetting and propose VDA, which preserves old-task VD through VC-OT and promotes visually grounded adaptation through VMA. Experiments on six tasks show that our proposed framework reduces forgetting while retaining strong plasticity.

\bibliography{aaai2027}

@article{chen2024coin,
  title={Coin: A benchmark of continual instruction tuning for multimodel large language models},
  author={Chen, Cheng and Zhu, Junchen and Luo, Xu and Shen, Heng T and Song, Jingkuan and Gao, Lianli},
  journal={Advances in neural information processing systems},
  volume={37},
  pages={57817--57840},
  year={2024}
}

@article{li2024llava,
  title={Llava-onevision: Easy visual task transfer},
  author={Li, Bo and Zhang, Yuanhan and Guo, Dong and Zhang, Renrui and Li, Feng and Zhang, Hao and Zhang, Kaichen and Zhang, Peiyuan and Li, Yanwei and Liu, Ziwei and others},
  journal={arXiv preprint arXiv:2408.03326},
  year={2024}
}

@misc{bai2025qwen25vltechnicalreport,
      title={Qwen2.5-VL Technical Report}, 
      author={Shuai Bai and Keqin Chen and Xuejing Liu and Jialin Wang and Wenbin Ge and Sibo Song and Kai Dang and Peng Wang and Shijie Wang and Jun Tang and Humen Zhong and Yuanzhi Zhu and Mingkun Yang and Zhaohai Li and Jianqiang Wan and Pengfei Wang and Wei Ding and Zheren Fu and Yiheng Xu and Jiabo Ye and Xi Zhang and Tianbao Xie and Zesen Cheng and Hang Zhang and Zhibo Yang and Haiyang Xu and Junyang Lin},
      year={2025},
      eprint={2502.13923},
      archivePrefix={arXiv},
      primaryClass={cs.CV},
      url={https://arxiv.org/abs/2502.13923}, 
}

@article{bai2025qwen3,
  title={Qwen3-vl technical report},
  author={Bai, Shuai and Cai, Yuxuan and Chen, Ruizhe and Chen, Keqin and Chen, Xionghui and Cheng, Zesen and Deng, Lianghao and Ding, Wei and Gao, Chang and Ge, Chunjiang and others},
  journal={arXiv preprint arXiv:2511.21631},
  year={2025}
}

@misc{wang2025internvl35advancingopensourcemultimodal,
      title={InternVL3.5: Advancing Open-Source Multimodal Models in Versatility, Reasoning, and Efficiency}, 
      author={Weiyun Wang and Zhangwei Gao and Lixin Gu and Hengjun Pu and Long Cui and Xingguang Wei and Zhaoyang Liu and Linglin Jing and Shenglong Ye and Jie Shao and Zhaokai Wang and Zhe Chen and Hongjie Zhang and Ganlin Yang and Haomin Wang and Qi Wei and Jinhui Yin and Wenhao Li and Erfei Cui and Guanzhou Chen and Zichen Ding and Changyao Tian and Zhenyu Wu and Jingjing Xie and Zehao Li and Bowen Yang and Yuchen Duan and Xuehui Wang and Zhi Hou and Haoran Hao and Tianyi Zhang and Songze Li and Xiangyu Zhao and Haodong Duan and Nianchen Deng and Bin Fu and Yinan He and Yi Wang and Conghui He and Botian Shi and Junjun He and Yingtong Xiong and Han Lv and Lijun Wu and Wenqi Shao and Kaipeng Zhang and Huipeng Deng and Biqing Qi and Jiaye Ge and Qipeng Guo and Wenwei Zhang and Songyang Zhang and Maosong Cao and Junyao Lin and Kexian Tang and Jianfei Gao and Haian Huang and Yuzhe Gu and Chengqi Lyu and Huanze Tang and Rui Wang and Haijun Lv and Wanli Ouyang and Limin Wang and Min Dou and Xizhou Zhu and Tong Lu and Dahua Lin and Jifeng Dai and Weijie Su and Bowen Zhou and Kai Chen and Yu Qiao and Wenhai Wang and Gen Luo},
      year={2025},
      eprint={2508.18265},
      archivePrefix={arXiv},
      primaryClass={cs.CV},
      url={https://arxiv.org/abs/2508.18265}, 
}

@article{chen2025sefe,
  title={Sefe: Superficial and essential forgetting eliminator for multimodal continual instruction tuning},
  author={Chen, Jinpeng and Cong, Runmin and Zhao, Yuzhi and Yang, Hongzheng and Hu, Guangneng and Ip, Horace Ho Shing and Kwong, Sam},
  journal={arXiv preprint arXiv:2505.02486},
  year={2025}
}

@inproceedings{zhang2023vqacl,
  title={Vqacl: A novel visual question answering continual learning setting},
  author={Zhang, Xi and Zhang, Feifei and Xu, Changsheng},
  booktitle={Proceedings of the IEEE/CVF Conference on Computer Vision and Pattern Recognition},
  pages={19102--19112},
  year={2023}
}

@inproceedings{singh2026ttrv,
  title={Ttrv: Test-time reinforcement learning for vision language models},
  author={Singh, Akshit and Marjit, Shyam and Lin, Wei and Gavrikov, Paul and Yeung-Levy, Serena and Kuehne, Hilde and Feris, Rogerio and Doveh, Sivan and Glass, James and Mirza, M Jehanzeb},
  booktitle={Proceedings of the IEEE/CVF Conference on Computer Vision and Pattern Recognition},
  pages={33153--33163},
  year={2026}
}

@article{kirkpatrick2017overcoming,
  title={Overcoming catastrophic forgetting in neural networks},
  author={Kirkpatrick, James and Pascanu, Razvan and Rabinowitz, Neil and Veness, Joel and Desjardins, Guillaume and Rusu, Andrei A and Milan, Kieran and Quan, John and Ramalho, Tiago and Grabska-Barwinska, Agnieszka and others},
  journal={Proceedings of the national academy of sciences},
  volume={114},
  number={13},
  pages={3521--3526},
  year={2017},
  publisher={National Academy of Sciences}
}

@article{li2017learning,
  title={Learning without forgetting},
  author={Li, Zhizhong and Hoiem, Derek},
  journal={IEEE transactions on pattern analysis and machine intelligence},
  volume={40},
  number={12},
  pages={2935--2947},
  year={2017},
  publisher={IEEE}
}

@inproceedings{rebuffi2017icarl,
  title={icarl: Incremental classifier and representation learning},
  author={Rebuffi, Sylvestre-Alvise and Kolesnikov, Alexander and Sperl, Georg and Lampert, Christoph H},
  booktitle={Proceedings of the IEEE conference on Computer Vision and Pattern Recognition},
  pages={2001--2010},
  year={2017}
}

@article{lopez2017gradient,
  title={Gradient episodic memory for continual learning},
  author={Lopez-Paz, David and Ranzato, Marc'Aurelio},
  journal={Advances in neural information processing systems},
  volume={30},
  year={2017}
}

@article{mallya2017packnet,
  title={Packnet: Adding multiple tasks to a single network by iterative pruning},
  author={Mallya, Arun and Lazebnik, Svetlana},
  journal={arXiv preprint arXiv:1711.05769},
  year={2017}
}

@inproceedings{qian2023decouple,
  title={Decouple before interact: Multi-modal prompt learning for continual visual question answering},
  author={Qian, Zi and Wang, Xin and Duan, Xuguang and Qin, Pengda and Li, Yuhong and Zhu, Wenwu},
  booktitle={Proceedings of the IEEE/CVF International Conference on Computer Vision},
  pages={2953--2962},
  year={2023}
}

@article{he2026continual,
  title={Continual instruction tuning for large multimodal models},
  author={He, Jinghan and Guo, Haiyun and Zhu, Kuan and Tang, Ming and Wang, Jinqiao},
  journal={IEEE Transactions on Image Processing},
  year={2026},
  publisher={IEEE}
}

@article{madaan2021representational,
  title={Representational continuity for unsupervised continual learning},
  author={Madaan, Divyam and Yoon, Jaehong and Li, Yuanchun and Liu, Yunxin and Hwang, Sung Ju},
  journal={arXiv preprint arXiv:2110.06976},
  year={2021}
}

@inproceedings{huai2025cl,
  title={Cl-moe: Enhancing multimodal large language model with dual momentum mixture-of-experts for continual visual question answering},
  author={Huai, Tianyu and Zhou, Jie and Wu, Xingjiao and Chen, Qin and Bai, Qingchun and Zhou, Ze and He, Liang},
  booktitle={Proceedings of the computer vision and pattern recognition conference},
  pages={19608--19617},
  year={2025}
}

@inproceedings{guo2025hide,
  title={Hide-llava: Hierarchical decoupling for continual instruction tuning of multimodal large language model},
  author={Guo, Haiyang and Zeng, Fanhu and Xiang, Ziwei and Zhu, Fei and Wang, Da-Han and Zhang, Xu-Yao and Liu, Cheng-Lin},
  booktitle={Proceedings of the 63rd Annual Meeting of the Association for Computational Linguistics (Volume 1: Long Papers)},
  pages={13572--13586},
  year={2025}
}

@inproceedings{zhao2024lova3,
  title={Lova3: Learning to visual question answering, asking and assessment},
  author={Zhao, Hengyuan and Zhou, Pan and Gao, Difei and Bai, Zechen and Shou, Mike Zheng},
  booktitle={The Thirty-eighth Annual Conference on Neural Information Processing Systems},
  year={2024}
}

@inproceedings{huang2023large,
  title={Large language models can self-improve},
  author={Huang, Jiaxin and Gu, Shixiang and Hou, Le and Wu, Yuexin and Wang, Xuezhi and Yu, Hongkun and Han, Jiawei},
  booktitle={Proceedings of the 2023 conference on empirical methods in natural language processing},
  pages={1051--1068},
  year={2023}
}

@article{prasad2024self,
  title={Self-consistency preference optimization},
  author={Prasad, Archiki and Yuan, Weizhe and Pang, Richard Yuanzhe and Xu, Jing and Fazel-Zarandi, Maryam and Bansal, Mohit and Sukhbaatar, Sainbayar and Weston, Jason and Yu, Jane},
  journal={arXiv preprint arXiv:2411.04109},
  year={2024}
}

@article{zhang2026right,
  title={Right question is already half the answer: Fully unsupervised llm reasoning incentivization},
  author={Zhang, Qingyang and Wu, Haitao and Zhang, Changqing and Zhao, Peilin and Bian, Yatao},
  journal={Advances in neural information processing systems},
  volume={38},
  pages={67345--67372},
  year={2026}
}

@article{agarwal2026unreasonable,
  title={The unreasonable effectiveness of entropy minimization in llm reasoning},
  author={Agarwal, Shivam and Zhang, Zimin and Yuan, Lifan and Han, Jiawei and Peng, Hao},
  journal={Advances in Neural Information Processing Systems},
  volume={38},
  pages={107150--107180},
  year={2026}
}

@article{hu2025test,
  title={Test-time learning for large language models},
  author={Hu, Jinwu and Zhang, Zhitian and Chen, Guohao and Wen, Xutao and Shuai, Chao and Luo, Wei and Xiao, Bin and Li, Yuanqing and Tan, Mingkui},
  journal={arXiv preprint arXiv:2505.20633},
  year={2025}
}

@article{zhu2024self,
  title={Self-supervised visual preference alignment},
  author={Zhu, Ke and Ge, Zheng and Zhao, Liang and Zhang, Xiangyu},
  journal={arXiv preprint arXiv:2404.10501},
  year={2024}
}

@article{zhou2024calibrated,
  title={Calibrated self-rewarding vision language models},
  author={Zhou, Yiyang and Fan, Zhiyuan and Cheng, Dongjie and Yang, Sihan and Chen, Zhaorun and Cui, Chenhang and Wang, Xiyao and Li, Yun and Zhang, Linjun and Yao, Huaxiu},
  journal={Advances in Neural Information Processing Systems},
  volume={37},
  pages={51503--51531},
  year={2024}
}

@article{wei2026first,
  title={First SFT, second RL, third UPT: Continual improving multi-modal LLM reasoning via unsupervised post-training},
  author={Wei, Lai and Li, Yuting and Wang, Chen and Wang, Yue and Kong, Linghe and Huang, Weiran and Sun, Lichao},
  journal={Advances in Neural Information Processing Systems},
  volume={38},
  pages={62293--62318},
  year={2026}
}

@article{yu2026stabilizing,
  title={Stabilizing Unsupervised Self-Evolution of MLLMs via Continuous Softened Retracing reSampling},
  author={Yu, Yunyao and Wu, Zhengxian and Chen, Zhuohong and Xu, Hangrui and Liao, Zirui and Deng, Xiangwen and Liu, Zhifang and Shi, Senyuan and Wang, Haoqian},
  journal={arXiv preprint arXiv:2604.03647},
  year={2026}
}

@article{ye2026not,
  title={Not All Tokens See Equally: Perception-Grounded Policy Optimization for Large Vision-Language Models},
  author={Ye, Zekai and Li, Qiming and Feng, Xiaocheng and Chen, Ruihan and Li, Ziming and Ren, Haoyu and Chen, Kun and Tu, Dandan and Qin, Bing},
  journal={arXiv preprint arXiv:2604.01840},
  year={2026}
}

@article{parisi2018continual,
  title={Continual lifelong learning with neural networks: A review},
  author={Parisi, German I and Kemker, Ronald and Part, Jose L and Kanan, Christopher and Wermter, Stefan},
  journal={arXiv preprint arXiv:1802.07569},
  volume={990},
  year={2018},
  publisher={eprint}
}

@article{de2021continual,
  title={A continual learning survey: Defying forgetting in classification tasks},
  author={De Lange, Matthias and Aljundi, Rahaf and Masana, Marc and Parisot, Sarah and Jia, Xu and Leonardis, Ale{\v{s}} and Slabaugh, Gregory and Tuytelaars, Tinne},
  journal={IEEE transactions on pattern analysis and machine intelligence},
  volume={44},
  number={7},
  pages={3366--3385},
  year={2021},
  publisher={IEEE}
}

@inproceedings{singh2019towards,
  title={Towards vqa models that can read},
  author={Singh, Amanpreet and Natarajan, Vivek and Shah, Meet and Jiang, Yu and Chen, Xinlei and Batra, Dhruv and Parikh, Devi and Rohrbach, Marcus},
  booktitle={Proceedings of the IEEE/CVF conference on computer vision and pattern recognition},
  pages={8317--8326},
  year={2019}
}

@inproceedings{borisova2025scivqa,
  title={SciVQA 2025: Overview of the first scientific visual question answering shared task},
  author={Borisova, Ekaterina and Rauscher, Nikolas and Rehm, Georg},
  booktitle={Proceedings of the Fifth Workshop on Scholarly Document Processing (SDP 2025)},
  pages={182--210},
  year={2025}
}

@article{zhao2025mllm,
  title={Mllm-cl: Continual learning for multimodal large language models},
  author={Zhao, Hongbo and Zhu, Fei and Guo, Haiyang and Wang, Meng and Wang, Rundong and Meng, Gaofeng and Zhang, Zhaoxiang},
  journal={arXiv preprint arXiv:2506.05453},
  year={2025}
}

@inproceedings{hudson2019gqa,
  title={Gqa: A new dataset for real-world visual reasoning and compositional question answering},
  author={Hudson, Drew A and Manning, Christopher D},
  booktitle={Proceedings of the IEEE/CVF conference on computer vision and pattern recognition},
  pages={6700--6709},
  year={2019}
}

@inproceedings{sima2024drivelm,
  title={Drivelm: Driving with graph visual question answering},
  author={Sima, Chonghao and Renz, Katrin and Chitta, Kashyap and Chen, Li and Zhang, Hanxue and Xie, Chengen and Bei{\ss}wenger, Jens and Luo, Ping and Geiger, Andreas and Li, Hongyang},
  booktitle={European conference on computer vision},
  pages={256--274},
  year={2024},
  organization={Springer}
}

@article{zhang2023pmc,
  title={Pmc-vqa: Visual instruction tuning for medical visual question answering},
  author={Zhang, Xiaoman and Wu, Chaoyi and Zhao, Ziheng and Lin, Weixiong and Zhang, Ya and Wang, Yanfeng and Xie, Weidi},
  journal={arXiv preprint arXiv:2305.10415},
  year={2023}
}

@article{li2025multimodal,
  title={Multimodal Continual Instruction Tuning with Dynamic Gradient Guidance},
  author={Li, Songze and Gao, Mingyu and Su, Tonghua and Zhang, Xu-Yao and Wang, Zhongjie},
  journal={arXiv preprint arXiv:2511.15164},
  year={2025}
}

@inproceedings{he2024seekr,
  title={Seekr: Selective attention-guided knowledge retention for continual learning of large language models},
  author={He, Jinghan and Guo, Haiyun and Zhu, Kuan and Zhao, Zihan and Tang, Ming and Wang, Jinqiao},
  booktitle={Proceedings of the 2024 Conference on Empirical Methods in Natural Language Processing},
  pages={3254--3266},
  year={2024}
}

@inproceedings{wang2022dualprompt,
  title={Dualprompt: Complementary prompting for rehearsal-free continual learning},
  author={Wang, Zifeng and Zhang, Zizhao and Ebrahimi, Sayna and Sun, Ruoxi and Zhang, Han and Lee, Chen-Yu and Ren, Xiaoqi and Su, Guolong and Perot, Vincent and Dy, Jennifer and others},
  booktitle={European conference on computer vision},
  pages={631--648},
  year={2022},
  organization={Springer}
}

@inproceedings{smith2023coda,
  title={Coda-prompt: Continual decomposed attention-based prompting for rehearsal-free continual learning},
  author={Smith, James Seale and Karlinsky, Leonid and Gutta, Vyshnavi and Cascante-Bonilla, Paola and Kim, Donghyun and Arbelle, Assaf and Panda, Rameswar and Feris, Rogerio and Kira, Zsolt},
  booktitle={Proceedings of the IEEE/CVF conference on computer vision and pattern recognition},
  pages={11909--11919},
  year={2023}
}

@misc{peng2026deeperthoughtweakeraim,
      title={Deeper Thought, Weaker Aim: Understanding and Mitigating Perceptual Impairment during Reasoning in Multimodal Large Language Models}, 
      author={Ruiying Peng and Xueyu Wu and Jing Lei and Lu Hou and Yuanzheng Ma and Xiaohui Li},
      year={2026},
      eprint={2603.14184},
      archivePrefix={arXiv},
      primaryClass={cs.CV},
      url={https://arxiv.org/abs/2603.14184}, 
}

@inproceedings{khayatkhoei2025mllms,
  title={Mllms know where to look: Training-free perception of small visual details with multimodal llms},
  author={Khayatkhoei, Mahyar and Chhikara, Prateek and Ilievski, Filip and others},
  booktitle={International Conference on Learning Representations},
  volume={2025},
  pages={68194--68213},
  year={2025}
}

@article{jung2020continual,
  title={Continual learning with node-importance based adaptive group sparse regularization},
  author={Jung, Sangwon and Ahn, Hongjoon and Cha, Sungmin and Moon, Taesup},
  journal={Advances in neural information processing systems},
  volume={33},
  pages={3647--3658},
  year={2020}
}

@article{zhu2026sample,
  title={Sample-Aware Knowledge Association and Enhancement for Open-Vocabulary Continual Learning},
  author={Zhu, Zhilin and Ma, Zhiheng and Wang, Yabin and Song, Yaguang and Wang, Yaowei and Hong, Xiaopeng},
  journal={International Journal of Computer Vision},
  volume={134},
  number={7},
  pages={332},
  year={2026},
  publisher={Springer}
}

@inproceedings{zhu2024reshaping,
  title={Reshaping the online data buffering and organizing mechanism for continual test-time adaptation},
  author={Zhu, Zhilin and Hong, Xiaopeng and Ma, Zhiheng and Zhuang, Weijun and Ma, Yaohui and Dai, Yong and Wang, Yaowei},
  booktitle={European conference on computer vision},
  pages={415--433},
  year={2024},
  organization={Springer}
}

@article{zhu2026dance,
  title={Dance Across Shifts: Forward-Facilitation Continual Test-Time Adaptation through Dynamic Style Bridging},
  author={Zhu, Zhilin and Wang, Yabin and Ma, Zhiheng and Song, Yaguang and Wang, Yaowei and Hong, Xiaopeng},
  journal={arXiv preprint arXiv:2605.18608},
  year={2026}
}

@article{shao2026agentpatch,
  title={AgentPatch: Coarse-to-Fine Weak-Task Repair for Merging Agentic Multimodal Large Language Models},
  author={Shao, Zibo and Xiong, Baochen and Xu, Chengdong and Xiao, Linhui and Li, Kaichen and Gong, Haoran and Li, Yan and Song, Yaguang and Yang, Xiaoshan},
  journal={arXiv preprint arXiv:2608.06699},
  year={2026}
}

@article{shao2026pivotmerge,
  title={PivotMerge: Bridging Heterogeneous Multimodal Pre-training via Post-Alignment Model Merging},
  author={Shao, Zibo and Xiong, Baochen and Yang, Xiaoshan and Song, Yaguang and Zhang, Qimeng and Chen, Haifeng and Xu, Changsheng},
  journal={arXiv preprint arXiv:2604.22823},
  year={2026}
}

@article{xu2026evomas,
  title={EvoMAS: Learning Execution-Time Workflows for Multi-Agent Systems},
  author={Xu, Chengdong and Ke, Kaiqiang and Liu, Ziheng and Wei, Jiaqi and Shao, Zibo and Guo, Weile and Yu, Chao},
  journal={arXiv preprint arXiv:2605.08769},
  year={2026}
}

\end{document}